# Loglinear models for first-order probabilistic reasoning

James Cussens
Department of Computer Science
University of York
Heslington, York, YO10 5DD, UK
jc@cs.york.ac.uk

## Abstract

Recent work on loglinear models in probabilistic constraint logic programming is applied to first-order probabilistic reasoning. Probabilities are defined directly on the *proofs* of atomic formulae, and by marginalisation on the atomic formulae themselves. We use Stochastic Logic Programs (SLPs) composed of labelled and unlabelled definite clauses to define the proof probabilities. We have a conservative extension of first-order reasoning, so that, for example, there is a one-one mapping between logical and random variables. We show how, in this framework, Inductive Logic Programming (ILP) can be used to induce the features of a loglinear model from data. We also compare the presented framework with other approaches to first-order probabilistic reasoning.

## 1 Introduction

A framework which merges first-order logical and probabilistic inference in a theoretically sound and applicable manner promises many benefits. We can benefit from the compact knowledge representation of logic, and still represent and reason about the uncertainty found in most applications. Here we propose a conservative extension to the logic programming framework by defining probabilities directly on proofs and hence indirectly on atomic formulae. Our conservatism allows us to tie probabilistic and logical concepts very closely. Table 1 lists the linkages which the proposed approach establishes.

This paper is laid out as follows. We begin in Section 2, with a brief overview of logic programming concepts. Section 3 forms the core of the paper where we introduce the loglinear model and Stochastic Logic Programs. Section 4 then presents SLPs which represent Markov nets and then more complex models. Section 5 discusses the role of ILP in learning the structure of the loglinear model, focusing on the work of Dehaspe. We discuss related work in Section 6 and briefly mention future work in Section 7.

| Logic | Probability |
|---|---|
| logical variable | random variable |
| instantiation | instantiation |
| relations | joint distributions |
| queries | queries |
| ground definitions | probability tables |
| disjunctive definitions | mixture models |
| defining relations in terms of other relations | defining distributions in terms of other distributions |

Table 1: Linking logic and probability

## 2 Logic programming essentials

We give a very brief overview of logic programming. For more details, the reader can consult any standard textbook on logic programming, e.g. (Lloyd, 1987). In this paper we will consider only definite logic programs. *Definite (logic) programs* consist of a set of definite clauses, where each *definite clause* is a disjunctive first-order formula such as $p(X,Y) \vee \neg q(X,Z) \vee \neg r(Z) \Leftrightarrow p(X,Y) \leftarrow q(X,Z) \wedge r(Z)$. All variables are implicitly universally quantified (we will denote variables by names starting with upper-case letters). A *literal* is an atomic formula (briefly *atom*) or the negation of an atom. Definite clauses consist of exactly one positive literal ($p(X,Y)$) in our example) and zero or more negative literals (such as $q(X,Z)$ and $r(Z)$). The positive literal is the *head* of the clause and the negative literals are the *body*.

A *goal* or *query* is of the form $\leftarrow Atom_1 \wedge Atom_2 \wedge \cdots \wedge Atom_n$. A *substitution*, such as $\theta = \{X/a, Y/Z\}$ is a mapping from variables to first-order terms. If a substitution maps variables to terms which do not include vari-ables, we will call it an *instantiation*. A substitution $\theta$ *unifies* two atoms $Atom_1$, $Atom_2$ if $Atom_1\theta$ ($\theta$ applied to $Atom_1$) is identical to $Atom_2\theta$. *Resolution* is an inference rule that takes an atom $Atom$ selected from a goal $\leftarrow Atom_1 \land \cdots \land Atom \land \cdots \land Atom_n$, unifies $Atom$ with the head $H$ of a clause $H \leftarrow B$ using a substitution $\theta$ and returns $(\leftarrow Atom_1 \land \cdots \land B \land \cdots \land Atom_n)\theta$ as a new goal. Note that $B$ may be empty. With Prolog the selected atom is always the leftmost atom. An *SLD-refutation* of a goal $G$ is a sequence of resolution steps which produce the empty goal. The *SLD-tree* for a goal $G$ is a tree of goals, with $G$ as root node, and such that the children of any node/goal $G'$ are goals produced by one resolution step using $G'$ (the empty goal has no children). Branches of the SLD-tree ending in the empty goal are *success branches* corresponding to successful refutations. The *success set* for a definite program is the set of all ground atoms $Atom$ such that $\leftarrow Atom$ has an SLD-refutation. The *success set for an n-arity predicate $p/n$*, denoted $SS(p/n)$, is all those atoms in the program's success set that have $p/n$ as their predicate symbol. *The most general goal* for a predicate $p/n$ is of the form $\leftarrow p(X_1, X_2, \ldots, X_n)$ where the $X_i$ are distinct variables. The *computed answer* for a goal is a substitution for the variables in $G$ produced by an SLD-refutation of $G$. We will sometimes use Prolog notation, where $p(X,Y) \leftarrow q(X,Z) \land r(Z)$ is represented thus:
`p(X,Y) :- q(X,Z), r(Z).`,
and $\leftarrow q(X,a)$ is represented thus:
`:- q(X,a).`

## 3 Loglinear models for first-order probabilistic reasoning

A loglinear probability distribution on a set $\Omega$ is of the following form. For $\omega \in \Omega$:

$$p(\omega) = Z^{-1} \exp\left(\sum_i \lambda_i f_i(\omega)\right) \quad (1)$$

where the $f_i$ are the features of the distribution, the $\lambda_i$ are the model parameters and $Z$ is a normalising constant.

### 3.1 Probabilistic Constraint Logic Programming

(Riezler, 1997) develops (Abney, 1997) by defining a loglinear model on the *proofs* of formulae with some constraint logic program. This requires defining features on these proofs (the $f_i$) and defining the model parameters (the $\lambda_i$).

The essentials of this approach can be given by using the logic programming framework. This is a special case of constraint logic programming where the only constraints allowed are equational constraints between terms. We will stay with the standard logic programming framework for simplicity. Consider the logic program $LP1$ given in Fig 1.

```
s(X) :- p(X,Y), q(Y).
p(a,b).  p(a,a).  p(a,c).  p(d,b).
q(b).  q(c).
```

Figure 1: $LP1$, a simple logic program

Fig 2 shows the SLD-tree generated by the query $\leftarrow s(X)$ (with empty goals omitted). This shows three refutations of $\leftarrow s(X)$ which amount to two proofs of $s(a)$ and one proof of $s(d)$.

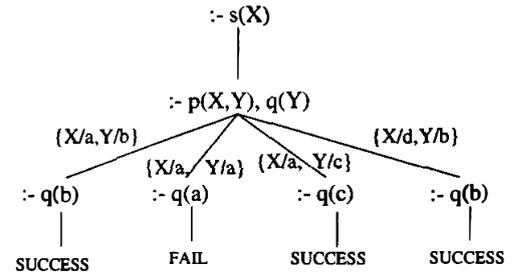

Figure 2: SLD-tree for $\leftarrow s(X)$ and $LP1$

We can now define a loglinear distribution on refutations of $\leftarrow s(X)$. Firstly, we define the features of the distribution. Consider two features of refutations, $f_1$ and $f_2$. For any refutation $R$, $f_1(R) = n$ if the goal $\leftarrow q(b)$ appears $n$ times in the $R$, and $f_2(R) = m$ if $\leftarrow q(c)$ appears $m$ times in $R$. Let $\lambda_1 = 0.2$ and $\lambda_2 = 0.4$, then the leftmost proof of $p(a)$ has probability $Z^{-1} \exp(0.2 \times 1 + 0.4 \times 0)$ and the one further to the right has probability $Z^{-1} \exp(0.2 \times 0 + 0.4 \times 1)$. The probability of the single proof of $p(d)$ is $Z^{-1} \exp(0.2 \times 1 + 0.4 \times 0)$. $Z$ is simply $\exp(0.2 \times 1 + 0.4 \times 0) + \exp(0.2 \times 0 + 0.4 \times 1) + \exp(0.2 \times 1 + 0.4 \times 0) = 2e^{0.2} + e^{0.4} \approx 3.9$, so the three probabilities are $0.31, 0.38, 0.31$ respectively. Having defined probabilities $p$ on the proofs of these atomic formulae, it is now trivial to define a distribution $p'$ on the formulae themselves: $p'(Atom) = \sum_{R \text{ is a proof of } Atom} p(R)$. We have $p'(s(a)) = 0.69$ and $p'(s(b)) = 0.31$, which is a distribution on $SS(s/1)$, the success set for $s/1$.

This approach applies very naturally to natural language processing (NLP). In NLP, a proof that a sentence belongs to a language amounts to a parse of that sentence, and the loglinear model can be used to find the most likely parse of any particular sentence. Riezler extends the improved iterative scaling algorithm of (Pietra et al., 1997) to induce features and parameters for a loglinear model from incomplete data. Incomplete data here consists of just atoms, rather than the proofs of those atoms. In an NLP context this means having a corpus of sentences rather than sentences annotated with their correct parses, the former being a considerably cheaper resource.

## 3.2 Stochastic Logic Programs

Riezler's framework allows arbitrary features of SLD-trees, and recent experiments have used features "indicating the number of argument-nodes or adjunct-nodes in the tree, and features indicating complexity, parallelism or branching-behaviour" (Stefan Riezler, personal communication).

Here we concentrate on a special case of Riezler's framework, where *the clauses used in a proof are the features defining the probability of that proof*, with clause labels denoting the parameters. (Eisele, 1994) examined this approach from an NLP perspective. (Muggleton, 1995) introduced Stochastic Logic Programs, approaching the issue from a general logic programming perspective, with a view to applications in Inductive Logic Programming.

In these cases, Stochastic Context-Free Grammars (SCFGs) were "lifted" to stochastic feature grammars (SFGs) and stochastic logic programs (SLPs) respectively. SCFGs are CFGs where each production is labelled, such that the labels for a particular non-terminal sum to one. The probability of a parse is then simply the product of the labels of all production rules used in that parse. Sentence probabilities are given by the sum of all parses of a sentence. The distributions so defined are special cases of loglinear models where the grammar rules define the features $f_i$ and their labels are the parameters $\lambda_i$. $Z$ is guaranteed to be one. This is because the labels for each non-terminal sum to one and because the context-freeness ensures that *we never fail*, and hence never have to backtrack, when generating a sentence from a SCFG—a production rule can always be applied to a nonterminal. Because of this a number of techniques (such as the inside-outside algorithm for parameter estimation (Lari and Young, 1990)) can be applied to SCFGs, but cannot be lifted to SFGs or SLPs. (See (Abney, 1997) for a demonstration of this.)

We define a stochastic logic program (SLP) as follows. An SLP is a logic program where some of the clauses are labelled with a non-negative number, and which satisfies the following constraints:

**Constraint 1** If there is a refutation of the most general goal for a predicate that uses a labelled clause, then the predicate is *distribution-defining*. It is required that the computed answer substitutions for any unit goal where the predicate is distribution-defining be ground.

**Constraint 2** The *potential* $\psi(R)$ of any refutation $R$ is the product of all the clause labels of the clauses used in $R$. If none of the clauses used in $R$ have labels, then $\psi(R)$ is undefined. The potential $\psi(G)$ of a goal is $\sum_{R \in ref(G)} \psi(R)$, where $ref(G)$ is the set of all refutations $R$ of $G$ such that $\psi(R)$ is defined. If $\psi(R)$ is undefined for all refutations $R$ of a goal $G$, then $\psi(G)$ is also undefined. We require that all goal potentials be finite.

**Constraint 3** For every distribution-defining predicate, the potential of its most general goal must be positive.

Constraint 1 can be met by requiring SLPs to be range-restricted: every variable appearing in the head of a clause must also appear in the body. The second condition is trivially met by any SLP where there is a bound on the depth of any refutation, e.g. non-recursive SLPs, and can also be met by requiring the clause labels for the clauses defining any given predicate to sum to at most one. Our definition generalises that found in (Muggleton, 1995), where Muggleton requires SLPs to be range-restricted and with labels for the same predicate summing exactly to one. Also, Muggleton does not use SLPs to define a loglinear model as we do here.

An SLP defines a distribution *for every distribution-defining predicate* in the SLP. Suppose $r/3$ were a distribution-defining predicate, then we have a loglinear distribution over refutations $R$ of the most general goal for this predicate $\leftarrow r(X_1, X_2, X_3)$, as follows:

$$p_r(R) = Z_r^{-1} \exp\left(\sum_i \log(\lambda_i) f(R, i)\right) \quad (2)$$

$$= Z_r^{-1} \prod_i \lambda_i^{f(R,i)} \quad (3)$$

where $\lambda_i$ is the label of clause $C_i$ and $f(R, i)$ is the number of times the clause $C_i$ is used in $R$. So we have a loglinear model where the labelled clauses define features and the logs of the clause labels are the model parameters.

$Z_r$ is simply the appropriate normalising constant, which can be found by simply summing the potentials of all refutations of $\leftarrow r(X_1, X_2, X_3)$. By definition, this sum is the potential of the goal $\leftarrow r(X_1, X_2, X_3)$. We have that $Z_r = \psi(\leftarrow r(A, B, C)) = \sum_{Atom \in SS(r/3)} \psi(\leftarrow Atom)$.

This last equation holds because Constraint 1 ensures that every refutation of $\leftarrow r(X_1, X_2, X_3)$ finds a member of $SS(r/3)$, and all elements of $SS(r/3)$ can be found this way. Constraints 2 and 3 ensure that $0 < Z_r < \infty$, so $Z_r^{-1}$ is always defined.

We get a marginal distribution over $SS(r/3)$: any ground atom $A$ has probability $p'_r(A) = \sum_{R \in ref(\leftarrow A)} p_r(R)$. Now consider the $X_i$ in $\leftarrow r(X_1, X_2, X_3)$. Each atom in $SS(r/3)$ defines a joint instantiation of the $X_i$ and therefore the distribution on atoms defines a three-dimensional joint probability distribution over $(D_1, D_2, D_3)$, where $D_i$ is the domain of $X_i$ which is both a logical and random variable. $D_i$ is just the set of values found for $X_i$ in $SS(r/3)$. In a standard logic program the $D_i$ will be finite or countably infinite.

We have used an example predicate $r/3$ to concreteness, but all of the above applies to predicates of any arity. We can use the logical structure of SLPs to define complex multi-dimensional joint distributions. The next section describes presents some SLPs, beginning with the simplest SLPs which represent Markov nets.

## 4 SLP models

### 4.1 SLPs and Markov nets

Markov nets (or undirected Bayes nets) are representations of *graphical* models, a special case of loglinear models. Fig 3 shows the "Asia" Markov net used as a running example in (Lauritzen and Spiegelhalter, 1988).

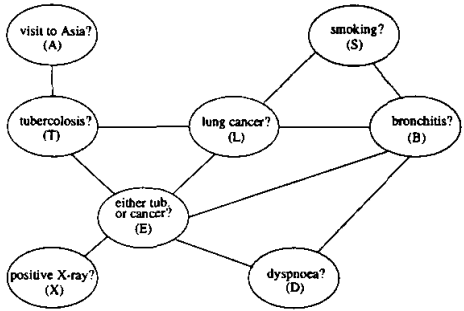

Figure 3: "Asia" Markov net

In general, let $V_C$ be the set of cliques of a Markov net $G$. A potential representation consists of *evidence potentials* $\psi_A$, defined on the cliques. Potentials are real-valued non-negative functions depending only on the states of the variables in each clique. The evidence potentials define a joint distribution on the nodes $V$ of the net as follows:

$$p(V) = Z^{-1} \prod_{A \in V_C} \psi_A \quad (4)$$

where

$$Z = \sum_V \prod_{A \in V_C} \psi_A \quad (5)$$

is a normalising constant. If $Z = 0$ then $p$ is undefined. ($Z$ will always be finite.) Consider the Markov net in Fig 3, which has six cliques. Each of the random variables in this net is binary, taking values $t$ or $f$. Table 2 gives a potential function defined on the clique $\{A, T\}$.

| Instantiation | | $\psi_{\{A,T\}}$ |
|---|---|---|
| $A = t$ | $T = t$ | 0.0005 |
| $A = t$ | $T = f$ | 0.0095 |
| $A = f$ | $T = t$ | 0.0099 |
| $A = f$ | $T = f$ | 0.9801 |

Table 2: An evidence potential on the clique $\{A, T\}$

Markov nets consist of a structure with associated parameters. Both can be represented easily using SLPs. Clique potentials are represented as tables of SLP ground facts. Figure 4 gives an SLP representation of the clique potential defined in Table 2.

```
0.0005 : c1(t,t).
0.0095 : c1(t,f).
0.0099 : c1(f,t)
0.9801 : c1(f,f).
```

Figure 4: Ground SLP representation of clique potential on $\{A, T\}$

We can then use a single unlabelled clause to represent the structure of a Markov net. The net in Fig 3 is represented by the unlabelled clause shown in Fig 5. Let us call SLPs that represent Markov nets in this fashion *Markov net SLPs*. Each ground goal has one refutation, so the probability of any ground atom is a normalised product; it is clear that the SLP and the Markov net represent the same loglinear distribution. Since they represent Markov nets, probabilistic inference based on such SLPs can use any of the standard algorithms for Markov nets.

```
asia(A,B,D,E,L,S,T,X) :-
   c6(E,X),
   c5(E,B,D),
   c4(L,B,X),
   c3(L,E,B),
   c2(E,L,T),
   c1(A,T).
```

Figure 5: Clausal representation of the "Asia" Markov net

### 4.2 SLP mixture models for context-specific independence

Consider the SLP in Figure 6 which represents a simple linear Markov net. We have that $A$ is independent of $C$ given $B$ ($A \perp C|B$). This conditional independence phenomenon is central to probabilistic graphical models such as Markov nets. But note that $A$ is independent of $C$ given *any* value of $B$. Sometimes we may not be justified in making such a strong assumption. It may be that $A$ is only independent of $C$ given *particular values of* $B$. This *conditional conditional independence*[1] or *context-specific independence* between $A$ and $C$ crops up often in applications and has been investigated by a number of workers, e.g. (Boutilier et al., 1996).

To represent context-sensitive independence, we need to be able to differentiate between these two sorts of values of $B$. Let us assume we have two predicates, strong/1 and weak/1 defined to be mutually exclusive which achieves

---
[1] conditional on a variable *and* conditional on values of that variable

```
linear(A,B,C) :-
    c1(A,B),
    c2(B,C).

%ground labelled definitions
%of c1 and c2 omitted
```

Figure 6: Linear Markov net SLP

this. We can then define the SLP in Fig 7 that defines an appropriate *mixture model*. A neater alternative might be to use negation to differentiate, using strong(B) and \+ strong(B)[2], but the use of negation in SLPs has yet to be properly investigated, hence our current restriction to definite clauses. Mixture models for context-specific independence are investigated in (Thiesson et al., 1997), where learning of such models is considered. (One can view tables defining discrete distributions as in Fig 4, as mixtures of degenerate distributions, but we will not do so.)

```
mixlin(A,B,C) :-
    strong(B),
    c1(A,B),
    c2(B,C).

mixlin(A,B,C) :-
    weak(B),
    c3(A,B,C).

%ground labelled definitions
%of c1, c2 and c3 omitted
```

Figure 7: Mixture model SLP defining context-specific independence

Context-sensitivity occurs whenever backtracking (due to unification failure) is a possibility in the search for refutations, and is ubiquitous in (real) natural language grammars. Fig 8 shows an SLP defining a distribution over the non-context-free language $\{a^n b^n c^n : n \geq 0\}$. Note that we can define distributions using structured terms, not just constants, and that the domain of this distribution is countably infinite.

```
anbncn(A) :- build(A-B,B-C,C-[]).

0.3: build(A-A,B-B,C-C).
0.7: build([a|A]-Ap,[b|B]-Bp,[c|C]-Cp) :-
     build(A-Ap,B-Bp,C-Cp).
```

Figure 8: Stochastic non context-free grammar defined with an SLP

---

[2]\+ is ISO Prolog notation for not.

### 4.3 Inference in SLP models

Markov nets, mixtures of Markov nets and context-sensitive stochastic grammars are all models that have been investigated in previous work, as have corresponding algorithms for inference and learning. Our aim here is to use SLPs as a common framework which can bring out useful connections and contrasts between these different models and algorithms.

A basic probabilistic inference problem in SLPs is to take a query, e.g. $\leftarrow t(X_1, a, X_3)$ and return $Pr_t(X_1, X_3 | X_2 = a)$, where $Pr_t$ is the distribution associated with the predicate $t/3$. The simple naive approach to inference in SLPs is to look for all refutations of $\leftarrow t(X_1, a, X_3)$, record their potentials and find $Pr_t(X_1, X_3 | X_2 = a)$ by marginalising. Although this could be used where we know that goals will have few refutations, in general it will be very inefficient and will not even terminate for goals with infinitely many refutations.

We do not have efficient general purpose algorithms for SLPs, so here we just sketch an approach. For a given query, find all clauses which have heads which unify with the goal, then apply the unifying substitution to the clause body, and then attempt to refute the subgoal composed of all the literals in the body that are *not* distribution-defining. For each clause body, and for each successful refutation, we have a remaining subgoal involving only distribution-defining predicates. Some of the variables in this remaining subgoal may be instantiated, so the subgoal represents a partially instantiated Markov net, but one where the functions defined on the cliques may not be represented by tables. When they are, we can use standard Markov net inference algorithms. When they are not, one possibility is to call our sketch algorithm recursively, if the SLP is so defined to guarantee termination. Note that we will only be interested in the distribution over the variables that appear in the head of the clause. These distributions can then be mixed according to the relevant clause labels to produce the final distribution.

## 5 Using ILP for feature construction

Since we use clauses to define the structural features of our distribution, it is natural to look to ILP for techniques which induce such structural features from data. (Dehaspe, 1997) does just this using the MACCENT algorithm which constructs a log-linear model using boolean clausal constraints as features. Dehaspe uses the "learning from interpretations" ILP setting where *each example is represented as a Prolog database*. Dehaspe applies MACCENT to classification, using a simple animal classification task to illustrate his approach. To bring out the connections between Dehaspe's approach and that presented here, we can rewrite Dehaspe's clausal constraints as labelled clauses as in Fig 9. Dehaspe uses negation which is safe here since it is assumed that all queries are ground.

```
L1 : p(X,fish) :-
   \+ has_legs(X), habitat(X,water).
L2 : p(X,reptile) :-
   \+ has_covering(X,hair), \+ has_legs(X).
```

Figure 9: Dehaspe's clausal constraints as labelled clauses

Dehaspe associates (modulo our rewrite) boolean features with each labelled clause, defined on $(I, Class)$ pairs, where $I$ denotes an, as yet, unclassified instance.

$$f_j(I, Class) = \begin{cases} 1 & \text{if } B, C_j \vdash p(I, Class) \\ 0 & \text{otherwise} \end{cases}$$

$B$ is background knowledge represented by a logic program. This defines a distribution over pairs $(I, Class)$,

$$Pr(I, Class) = Z^{-1} \exp\left(\sum_j \lambda_j f_j(I, Class)\right)$$

and hence, with suitable normalisation, conditional distributions $Pr(Class|I)$. We have a bijection between proofs of $p/2$ atoms and $p/2$ clauses, since each proof uses exactly one $p/2$ clause. This allows Dehaspe to treat *each proof as a feature*, where the parameter associated with each feature (=proof) is the label of the $p/2$ clause used in that proof. These features are then used to define a probability on atoms directly.

This contrasts with the SLP and PFG approach where each proof *has* features (e.g. the set of labelled clauses used in the proof), and these are used to define a *probability over proofs*. To get an (unnormalised) probability on an atom with SLPs we have to sum up the probabilities of the proofs of that atom.

Dehaspe's approach allows a more direct definition of a distribution over atoms, but relies on each proof passing through exactly one labelled clause. SLPs are not so restricted. Also with SLPs, the probability of an atom always increases with the number of proofs of that atom, which seems desirable. Following Dehaspe's approach this may not always be the case.

Dehaspe exploits the lattice structure of clauses and applies ILP techniques to guide the search for suitable constraints, searching for clauses with a general-to-specific beam search using the DLAB declarative bias language formalism. Dehaspe, like Riezler, keeps all the old parameters fixed when searching for the next constraint (= clause).

There are techniques for learning the structure of Bayes nets which start from an unconnected net and incrementally add arcs. Such techniques are strongly related to ILP searches (like Dehaspe's) where we start from a maximally general clause e.g. $p(X, Y, Z) \leftarrow$ and *refine* it by adding literals to the body until a 'best' (however defined) clause is found. $p(X, Y, Z) \leftarrow$ corresponds to an totally unconnected Markov net with three nodes. Refining this, to, say, $p(X, Y, Z) \leftarrow q(X, Y)$ corresponds adding an arc between the $X$ and $Y$ nodes. Further exploration of this connection may well yield valuable cross-fertilisation between ILP and Bayes net structure learning.

## 6  Related work

We do not give anything like a comprehensive survey of the work on connecting logic and probability that can be found in the UAI, philosophical, statistical and logical literature. Instead we will contrast the approach presented here with a few examples of particularly closely related work.

This translation of the clique functions of a Markov net to a generalised relational database is essentially the same as that of (Wong et al., 1995). Wong *et al* translate many of the graphical operations used with Markov nets to database operations: product distributions are constructed using joins, conditional distributions by projection, and marginals by database operations which mimic the standard approach in the Markov net literature. Wong *et al*'s argument is that since the operations required for effective use of Markov nets are defined on tables—for example, tables defining marginal probability distributions—one should use optimised methods developed by the database community for manipulating tables. The current work seeks to extend that of Wong *et al* by moving from a relational database setting to the logic programming setting.

In *Knowledge-based model construction (KBMC)* (Ngo and Haddaway, 1997; Koller and Pfeffer, 1997; Haddaway, 1999) first-order rules with associated probabilities are used to generate Bayesian networks for particular queries. As in SLD-resolution queries are matched to the heads of rules, but in KBMC this results in nodes representing ground facts being added to a growing (directed) Bayesian network. A *context* is defined using normal first-order rules, perhaps explicitly as a logic program (Ngo and Haddaway, 1997), which specifies logical conditions for labelled rules to be used. The ground facts are seen as boolean variables (either *true* or *false*). Once the Bayesian network is built it is then used to compute the probability that the query is true.

In KBMC, as in much of the work connecting logic and probability, parameterised first-order rules $\alpha : c(X) \leftarrow a(X)$ are connected to conditional probability statements such as $p(c(b)|a(b)) = \alpha$. Also the objective is to compute the probability that an atom is true. In the current paper, we focus on undirected representations, so that $\lambda : p(X, Y) \leftarrow q(X, Y), r(Y)$ forms part of the definition of a binary distribution associated with $p/2$ defined in terms of distributions associated with $q/2$ and $r/1$. We make *no* attempt to model causality.

Secondly, we do not use a labelled rule $\lambda : p(X, Y) \leftarrow q(X, Y), r(Y)$ to define the probability that some ground atom $p(a, b)$ is *true* as in KBMC, or to provide bounds on the probability that $p(a, b)$ is true as in (Shapiro, 1983; Ng and Subrahmanian, 1992). Instead, we have a binary distribution associated with $p(X, Y)$ which defines the probability of instantiations such as $\{X/a, Y/b\}$. In order to reason about the probability of the truth of atoms, we simply augment atoms by introducing an extra logical-random variable to represent the truth value of unaugmented atoms, and then treat this logical-random variable exactly as any other. This is in keeping with our conservative approach—if we are interested in the truth value of an atom as it varies across different "possible worlds"—then we model this variation in the standard way: with a random variable.

Consider
```
genotype(P,G) :- (0.5) parent(P,Q),
genotype(Q,G).
```
an example from (Koller and Pfeffer, 1997), where the "rule says that when a person's parent has a gene, the person will inherit it with probability 0.5". We would encode such "degree of belief" probability in an SLP with a boolean truth-value variable as in Fig 10.

```
genotype(P,G,T) :- parent(P,Q),
   genotype(Q,G,1), half(T).

0.5 : half(1).
0.5 : half(0).
```

Figure 10: Representing degree of belief with an extra variable

To find the probability that $genotype(bob, big\_ears)$ is true we are required to use the SLP to compute the probabilities of $genotype(bob, big\_ears, 1)$ and $genotype(bob, big\_ears, 0)$. (In fact, all we need are unnormalised potentials for these, which is simpler.) This amounts to demanding arguments (=proofs) for the truth of $genotype(bob, big\_ears)$ *and* for its falsity. We then effectively balance the strength of these proofs when deciding on the probability of truth.

Despite these differences in approach there are clear similarities between KBMC query-specific Bayes net construction and the query-specific exploration of an SLD-tree by Prolog which deserve further investigation.

Another approach to relational probabilistic reasoning are the relational Bayesian networks of (Jaeger, 1997). Here whole interpretations are the nodes of a Bayesian net. It is conceivable that such networks could be implemented as an SLP, using some suitable object-level representation of an interpretation, but it is likely that they would be unwieldy in practice. Since, at the end of the day, we are interested in the truth-values of atoms, it seems easier to deal with these directly, perhaps resorting to quite complex SLPs to model complex interactions between degrees of belief.

Finally, SLPs are very closely related to the stochastic (functional) programs of (Koller et al., 1997). Stochastic execution of the functional program defines a distribution over outputs of the program. As we have done here, Koller *et al* show how Bayesian nets and SCFGs can be represented in their richer formalism. They base their representation of directed Bayes nets on "the observation that each node in a Bayes net is a stochastic function of its parent's values." They also show how their formalism can exploit context-sensitive independence. *Unlike* the present paper, they also provide details of an efficient algorithm for probabilistic inference in their formalism, which mimics standard efficient algorithms for Bayesian networks. (Koller et al., 1997), does not discuss methods for inducing stochastic functional programs, but it seems highly likely that ILP techniques could be applied.

## 7  Open questions and future work

We have shown how various properties of SLPs (shared variables, multi-clause programs, unification failure and existential variables) correspond to various existing models (graphical models, mixture models, context-sensitive models and marginalisation) and argued that existing algorithms for these models can hence be used for inference and learning in SLPs. This work remains to be done. It is likely that suitable algorithms will mimic algorithms those used in Koller *et al*'s stochastic programs. Work on the implementation of randomised algorithms in logic programming is likely to be relevant too (Angelopoulos et al., 1998). We also expect techniques from logic programming and computational linguistics, such as Earley deduction and program transformation to be useful. For example, when learning the parameters of SLPs, Riezler's approach of storing proofs in a chart using Earley deduction makes a lot more sense than continually re-refuting goals.

Probabilistic inference and learning by Markov Chain Monte Carlo is also attractive for SLPs. For example, in a Gibbs sampling approach, all except one argument of a goal would be ground on each iteration. Such constrained goals generally have few refutations which might lead to an efficient method.

Finally, we hope that the current framework will stimulate further research into statistical ILP, and that such research will benefit from and contribute to related work on inducing models from data in computational linguistics and Bayesian networks.

**Acknowledgements**

Many thanks to Sara-Jayne Farmer for weeding out various errors and omissions. Thanks to Stephen Muggleton for useful discussions on the role of normalisation in SLPs and to Stefan Riezler for clarifying his method. Thanks also Gillian, Jane and Robert Higgins for putting up with me. Finally thanks to Luc de Raedt for encouraging me to investigate first-order Bayesian nets.

# References

Abney, S. (1997). Stochastic attribute-value grammars. *Computational Linguistics*, 23(4):597–618.

Angelopoulos, N., Pierro, A. D., and Wicklicky, H. (1998). Implementing randomised algorithms in constraint logic programming. In *Proc. of the Joint International Conference and Symposium on Logic Programming, JICSLP'98*, Manchester, UK. MIT Press.

Boutilier, C., Friedman, N., Goldszmidt, M., and Koller, D. (1996). Context-specific independence in Bayesian networks. In *Proceedings of the Twelfth Annual Conference on Uncertainty in Artificial Intelligence (UAI-96)*, pages 115–123, Portland, Oregon.

Dehaspe, L. (1997). Maximum entropy modeling with clausal constraints. In *Inductive Logic Programming: Proceedings of the 7th International Workshop (ILP-97). LNAI 1297*, pages 109–124. Springer.

Eisele, A. (1994). Towards probabilistic extensions of constraint-based grammars. Contribution to DYANA-2 Deliverable R1.2B, DYANA-2 project.

Haddaway, P. (1999). An overview of some recent developments in Bayesian problem solving techniques. *AI Magazine*.

Jaeger, M. (1997). Relational Bayesian networks. In *Proceedings of the Thirteenth Annual Conference on Uncertainty in Artificial Intelligence (UAI-97)*, pages 266–273, San Francisco, CA. Morgan Kaufmann Publishers.

Koller, D., McAllester, D., and Pfeffer, A. (1997). Effective Bayesian inference for stochastic programs. In *Proc. of the Fourteenth National Conference on Artificial Intelligence (AAAI-97)*, pages 740–747, Providence, Rhode Island, USA.

Koller, D. and Pfeffer, A. (1997). Learning probabilities for noisy first-order rules. In *Proceedings of the Fifteenth International Joint Conference on Artificial Intelligence (IJCAI-97)*, Nagoya, Japan.

Lari, K. and Young, S. (1990). The estimation of stochastic context-free grammars using the Inside-Outside algorithm. *Computer Speech and Language*, 4:35–56.

Lauritzen, S. and Spiegelhalter, D. (1988). Local computations with probabilities on graphical structures and their applications to expert systems. *Journal of the Royal Statistical Society A*, 50(2):157–224.

Lloyd, J. (1987). *Foundations of Logic Programming*. Springer, Berlin, second edition.

Muggleton, S. (1995). Stochastic logic programs. In De Raedt, L., editor, *Proceedings of the 5th International Workshop on Inductive Logic Programming*, page 29. Department of Computer Science, Katholieke Universiteit Leuven.

Ng, R. and Subrahmanian, V. (1992). Probabilistic logic programming. *Information and Computation*, 101(2):150–201.

Ngo, L. and Haddaway, P. (1997). Answering queries from context-sensitive probabilistic knowledge bases. *Theoretical Computer Science*, 171:147–171. Special issue on uncertainty in databases and deductive systems.

Pietra, S. D., Pietra, V. D., and Lafferty, J. (1997). Inducing features of random fields. *IEEE Transactions on Pattern Analysis and Machine Intelligence*, 19(4):380–393.

Riezler, S. (1997). Probabilistic constraint logic programming. Arbeitsberichte des SFB 340 Bericht Nr. 117, Universität Tübingen.

Shapiro, E. (1983). Logic programs with uncertainties: A tool for implementing rule-based systems. In *Proc. IJCAI-83*, pages 529–532.

Thiesson, B., Meek, C., Chickering, D., and Heckerman, D. (1997). Learning mixtures of DAG models. Technical Report MSR-TR-97-30, Microsoft Research. Revised May 1998.

Wong, S. K. M., Butz, C. J., and Xiang, Y. (1995). A method for implementing a probabilistic model as a relational database. In *Proceedings of the Eleventh Annual Conference on Uncertainty in Artificial Intelligence (UAI-95)*, pages 556–564, Montreal, Quebec, Canada.